\documentclass[letterpaper]{article} 
\usepackage{aaai2026}  
\usepackage{times}  
\usepackage{helvet}  
\usepackage{courier}  
\usepackage[hyphens]{url}  
\usepackage{graphicx} 
\usepackage{natbib}  
\usepackage{caption} 
\usepackage{algorithm}
\usepackage{algorithmic}

\usepackage{booktabs}       
\usepackage{amsfonts}       
\usepackage{nicefrac}       
\usepackage{microtype}      
\usepackage{xcolor}         
\usepackage{makecell}
\usepackage{multirow}
\usepackage{amsmath}
\usepackage{tcolorbox}
\usepackage{newfloat}
\usepackage{listings}
\DeclareCaptionStyle{ruled}{labelfont=normalfont,labelsep=colon,strut=off} 
\floatstyle{ruled}
\newfloat{listing}{tb}{lst}{}
\floatname{listing}{Listing}
\nocopyright 

\definecolor{rosepink}{RGB}{255, 102, 204}

\usepackage[hyphens]{url}
\usepackage{hyperref}
\title{
  \raisebox{-0.26\height}{\includegraphics[width=22pt]{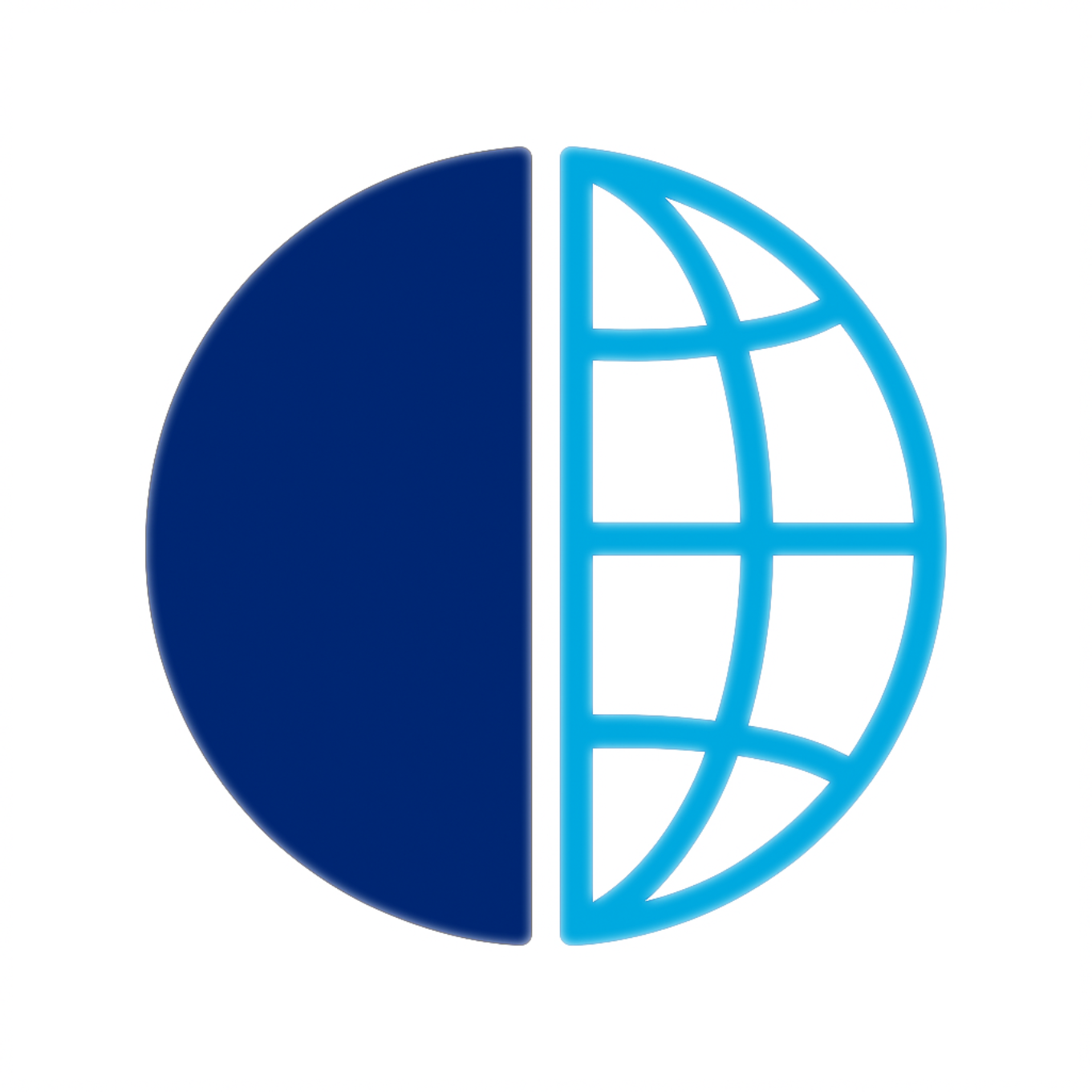}}MonoDream: Monocular Vision-Language Navigation with Panoramic Dreaming
}
\author{
    Shuo Wang\textsuperscript{\rm 1,3}\thanks{This work was done while Shuo Wang was a Research Intern with Horizon Robotics.}, Yongcai Wang\textsuperscript{\rm 1}\thanks{Corresponding authors},  Zhaoxin Fan\textsuperscript{\rm 2}\textsuperscript{\dag}, Yucheng Wang\textsuperscript{\rm 3}\thanks{Project leader}, Maiyue Chen\textsuperscript{\rm 3}, Kaihui Wang\textsuperscript{\rm 3}\\
    Zhizhong Su\textsuperscript{\rm 3}, Wanting Li\textsuperscript{\rm 1}, Xudong Cai\textsuperscript{\rm 1}, Yeying Jin\textsuperscript{\rm 4}, Deying Li\textsuperscript{\rm 1}
}
\affiliations{
    \textsuperscript{\rm 1}Renmin University of China\\
    \textsuperscript{\rm 2}Innovation Center for Future Blockchain and Privacy Computing, Beijing\\
    \textsuperscript{\rm 3}Horizon Robotics\\
    \textsuperscript{\rm 4}National University of Singapore\\ [2mm]
    \textcolor{rosepink}{\texttt{\url{https://horizonrobotics.github.io/robot\_lab/monodream}}}
}

\begin{document}

\maketitle

\begin{figure*}[ht]
    \centering
    \includegraphics[width=1\linewidth]{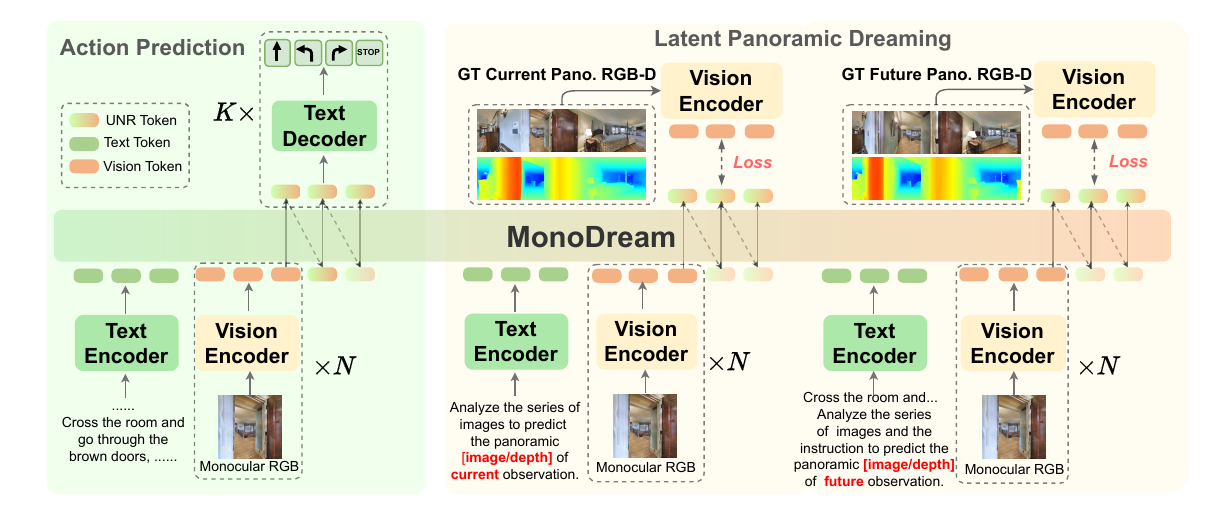}
    \caption{ Overview of the MonoDream framework.
MonoDream employs a Vision-Language Action (VLA) framework to encode both visual observations and textual instructions into a Unified Navigation Representation (UNR). The Action Prediction task generates the next action in natural language and is trained with action loss. The Latent Panoramic Dreaming (LPD) encourages the model to internally imagine the latent features of panoramic RGB-D images of current and future steps, providing global visual and geometric context via feature-basd loss. This multi-task co-training enables monocular agents to reason beyond the limited field of view and make more informed navigation decisions.}
    \label{fig:pipeline}
\end{figure*}

\begin{abstract}
Vision-Language Navigation (VLN) tasks often leverage panoramic RGB and depth inputs to provide rich spatial cues for action planning, but these sensors can be costly or less accessible in real-world deployments. Recent approaches based on Vision-Language Action (VLA) models achieve strong results with monocular input, yet they still lag behind methods using panoramic RGB-D information. We present MonoDream, a lightweight VLA framework that enables monocular agents to learn a Unified Navigation Representation (UNR). This shared feature representation jointly aligns navigation-relevant visual semantics (e.g., global layout, depth, and future cues) and language-grounded action intent, enabling more reliable action prediction. MonoDream further introduces Latent Panoramic Dreaming (LPD) tasks to supervise the UNR, which train the model to predict latent features of panoramic RGB and depth observations at both current and future steps based on only monocular input. Experiments on multiple VLN benchmarks show that MonoDream consistently improves monocular navigation performance and significantly narrows the gap with panoramic-based agents.
\end{abstract}


\section{Introduction}

Vision-Language Navigation tasks \cite{wu2024vision,anderson2018vision,gu2022vision} require embodied agents to follow language instructions and navigate to specified targets in 3D environments. Early successful approaches often rely on global perceptual inputs, such as panoramic RGB-D images \cite{hong2022bridging,wang2023gridmm,an2022bevbert}, which provide a wide field of view and explicit visual and geometric information. These inputs allow agents to build a more complete understanding of the environment and achieve high navigation success rates.

However, panoramic cameras and depth sensors introduce higher cost, power consumption, added weight and hardware integration complexity, making them less practical in many real-world deployments. Recent research has therefore focused on more lightweight settings where the agent is equipped with only a single forward-facing RGB camera \cite{zhang2024navid,zhang2024uni,cheng2024navila}. 

While monocular agents are easier to deploy, their navigation performance still lags significantly behind systems with panoramic RGB-D inputs. This gap arises from the narrow egocentric view of monocular agents, which limits the ability to infer latent global spatial and geometric cues that are beneficial for navigation. 
While such cues can be more directly captured from panoramic or depth observations, they are harder to extract from a monocular view.
For example, when navigating a building, the agent may see a long hallway but be unaware of a side door or staircase just outside its view. Such blind spots make it difficult to reason about local cues into a coherent global picture and plan several steps ahead, which are all critical for reliable navigation.

To tackle this issue, we propose \textbf{MonoDream}, a lightweight VLA framework that equips monocular VLN agents with a latent imagination capability. 
Our approach is built upon a key insight: navigation-relevant information, including the predicted actions and the understanding of the global scene, can be encoded into a shared representation space and could be inferred by the agent.
This insight is inspired by neuroscientific findings that the human brain reasons the current panoramic scene from partial views \cite{robertson2016neural} and internally simulates upcoming scenes based on intention \cite{seeber2025human}. MonoDream enables monocular agents to learn to infer and complete a holistic understanding of the current and future environments from limited egocentric observations.

To achieve this, MonoDream jointly aligns the navigation-relevant information, including implicit action intent, panoramic scene layout, depth perception, and future dynamics, into a unified latent space called the \textbf{Unified Navigation Representation (UNR)}. 
The UNR can be decoded into navigation actions or directly as features of global information.
We further design \textbf{Latent Panoramic Dreaming (LPD)} tasks that supervise the UNR by aligning the latent features of panoramic RGB-D observations at both the current and future steps. 
These tasks encourage the agent to develop a coherent, geometry-aware and future-aware internal model of the environment, enabling more informed navigation decisions from limited monocular input. Our contributions are as follows:

\begin{itemize}
    \item We propose \textbf{MonoDream}, a novel  monocular VLN framework that enhances the agent's internal global-aware ability. MonoDream enables the agent to infer implicit global, geometric and temporal context with monocular images.
    \item We introduce two key components: a Unified Navigation Representation that jointly encodes navigation actions and latent global scene, and Latent Panoramic Dreaming tasks that supervise UNR learning from current and future panoramic RGB-D latent features.
    \item We demonstrate the effectiveness of MonoDream by achieving state-of-the-art performance on the monocular VLN-CE benchmark, including R2R-CE and RxR-CE, while using a smaller VLA model without external training data. Cross-dataset evaluations further validate the strong generalization ability of our approach.
\end{itemize}

\section{Related Work}

\subsection{Vision Language Navigation}
Vision-Language Navigation tasks \cite{nguyen2019vision,wang2022towards,wu2020towards,schumann2024velma} challenge an embodied agent to follow natural language instructions and reach a target location within a previously unseen environment. Although early approaches largely focus on discrete navigation settings \cite{hong2021vln,chen2021history,liu2023bird}, where the agent moves along a predefined graph and typically has access to panoramic RGB-D observations, more recent studies \cite{hong2022bridging,wang2023gridmm,an2022bevbert,dai2024unitedvln} have shifted toward more realistic, continuous environments to predict low-level actions, and some works also begin to investigate monocular RGB-D settings.

Recent works based on large models \cite{liunavid,zhang2024uni,cheng2024navila,wei2025streamvln} leverage large-scale, RGB-only video models to build monocular VLN systems with enhanced generalization and real-world applicability.
However, agents in these works rely on a single forward-facing RGB camera, which poses additional challenges due to the limited field of view and the absence of explicit depth information.
Some works \cite{wang2024sim,wang2025dynam3d} attempt to use neural rendering techniques to recover observations from novel viewpoints from the reconstructed map. However, these methods rely on additional localization and mapping modules, and their performance still falls significantly short compared to approaches that directly utilize the panoramic input.

To tackle the limitation, MonoDream introduces LPD tasks, predicting the latent features of global and geometric information as auxiliary tasks, to enhance the performance of monocular VLN systems.
We demonstrate that our monocular navigation model with the unified navigation representation, equipped with panoramic, depth, and future awareness through auxiliary supervision, achieves state-of-the-art performance in monocular settings.

\subsection{Vision Imagination in Embodied Agents}
Learning to imagine vision information has become an increasingly popular strategy for improving policy learning in embodied agents, particularly when the agent operates under egocentric and limited monocular observations.
In such settings, inferring spatial layouts, geometric structures, and future dynamics beyond the current observation becomes essential for decision-making.
Early approaches~\cite{wu2023unleashing,wang2024reinforcement} often adopt a two-stage pipeline, where future states are predicted using off-the-shelf world models \cite{zheng2024open,bar2025navigation} or neural rendering \cite{mildenhall2021nerf,kerbl20233d} from reconstructed maps. 
Although simple and modular, these methods are inherently constrained by the accuracy and generalizability of the underlying world models.
More recent works \cite{cen2025worldvla,tian2024predictive,chen2024moto} advocate for end-to-end training paradigms that integrate forecasting and action planning within a unified framework. These methods allow the agent to jointly predict future observations and actions, leading to improved performance and generalization. Notably, DreamVLA \cite{zhang2025dreamvla} introduces a future-aware policy that forecasts dynamics, depth, and semantics simultaneously from monocular RGB input, demonstrating the benefits of future modeling in visual language tasks.

Our proposed MonoDream is the first VLN framework to learn a unified latent representation supervised by global panoramic scene layouts, while using only monocular RGB images as input. This panoramic-aware latent supervision equips the agent with a holistic spatial understanding essential for navigation, and leads to state-of-the-art performance under the monocular setting.

\section{Method}

\textbf{Overview.} 
We study monocular Vision-and-Language Navigation in Continuous Environments (VLN-CE), where an agent navigates in realistic spaces based on natural language instructions.
At each step $t$, the agent receives three types of inputs: the natural language instruction $\mathcal{I}$, the current egocentric RGB observation $o_t$, and a sampled history of past observations $\mathcal{O}_t$. 
The primary objective is to predict the next navigation action $a_t$ based on these multimodal inputs.

The framework of our proposed MonoDream (denoted as $\pi_\theta$ with the network parameters as $\theta$) is illustrated in Figure \ref{fig:pipeline}, which is built based on a Vision Language Model (VLM), including a vision encoder, a text encoder and decoder, and an LLM-based backbone.
MonoDream constructs the Unified Navigation Representation, a shared latent space produced by the VLM backbone, which is designed to jointly align navigation actions and the agent’s internal estimation of the implicit global information, including panoramic visual and structural cues for the current and future observations, based on only monocular inputs.
To supervise the formation of such representations, we introduce a set of auxiliary tasks termed Latent Panoramic Dreaming. These tasks are only applied during training, where the agent is guided to predict the latent features of:
(1) the panoramic RGB-D observation at the current step, and
(2) the panoramic RGB-D of a future step.
Through this joint design, MonoDream learns to align action decisions with imagined global and future context, thereby significantly enhancing monocular navigation performance.

\subsection{Unified Navigation Representation}

To enable the agent to internalize global and future scene awareness from limited monocular observations, we propose the Unified Navigation Representation that captures rich navigation-relevant information in a compact latent space.
UNR aligns navigation-relevant information together, including implicit features of actions, panoramic scene layout, panoramic depth perception, and future dynamics, into a shared latent space.
It can be decoded into navigation actions, instruction-like text, or directly as navigation-relevant features. 
In this section, we detail how language and visual observations are encoded into a shared feature space, which serves as the output of our MonoDream backbone.

For the input, the language instruction $\mathcal{I}$ is tokenized and processed by a text encoder, $\Phi_{\text{text}}(\cdot)$, to produce text feature $E_{\text{text}} \in \mathbb{R}^{L \times d}$, where $L$ is the text sequence length and $d$ is the hidden dimension.
\begin{equation}
    E_{\text{text}}  = \Phi_{\text{text}}(\mathcal{I})
\end{equation}

The agent's visual input contains the current view $o_t$ and a sampled historical images. To maintain computational efficient while preserving essential historical context, we uniformly sample $N$ frames, denoted as $\mathcal{O}_t = \{o_{p_0}, \dots, o_{p_{N-1}}\}$, from the full sequence of past observations. Each of these images is independently processed by the vision encoder, $\Phi_{\text{vis}}(\cdot)$, which encodes each image into $d$-dimensional feature and these individual image features are collected to form the visual input sequence $E_{\text{vis}}$. $E_{\text{vis}}$ represents the agent's complete visual context at step $t$ from monocular image sequence.

\begin{equation}
E_{\text{vis}} = \{ {\Phi_{\text{vis}}(o_{p}), \dots, \Phi_{\text{vis}}(o_{p_{N-1}}), \Phi_{\text{vis}}(o_t)} \}
\end{equation}

After encoding, the text feature $E_{\text{text}}$ and the vision feature $E_{\text{vis}}$ are combined into a single input sequence $S_t$ for the LLM-based backbone of MonoDream:
\begin{equation}
    S_t = [E_{\text{text}}, E_{\text{vis}}]
\end{equation}
This sequence is then fed into the backbone and the backbone outputs the hidden states $h_t \in \mathbb{R}^{l_{seq} \times d}$, where $l_{seq}$ is the length of the output sequence. This representation encapsulates the comprehensive state of the agent at step $t$:
\begin{equation}
    h_t = \text{MonoDream-Backbone}(S_t)
\end{equation}

We define hidden feature $h_t$ as the Unified Navigation Representation. To ensure that $h_t$ encodes all navigation-relevant information, even under limited monocular observations, we design a multitask training framework. In this framework, MonoDream learns $h_t$ to capture both vision-based signals (global layout, geometric structure, and future dynamics) and language-based signals (actions and instructions). Specifically, we jointly train the model with multiple objectives: Latent Panoramic Dreaming (Section \ref{sec:lpd}), action prediction and instruction reasoning (Section \ref{sec:cotrain}).

\begin{figure*}[htbp]
    \centering
    \includegraphics[width=0.83\linewidth]{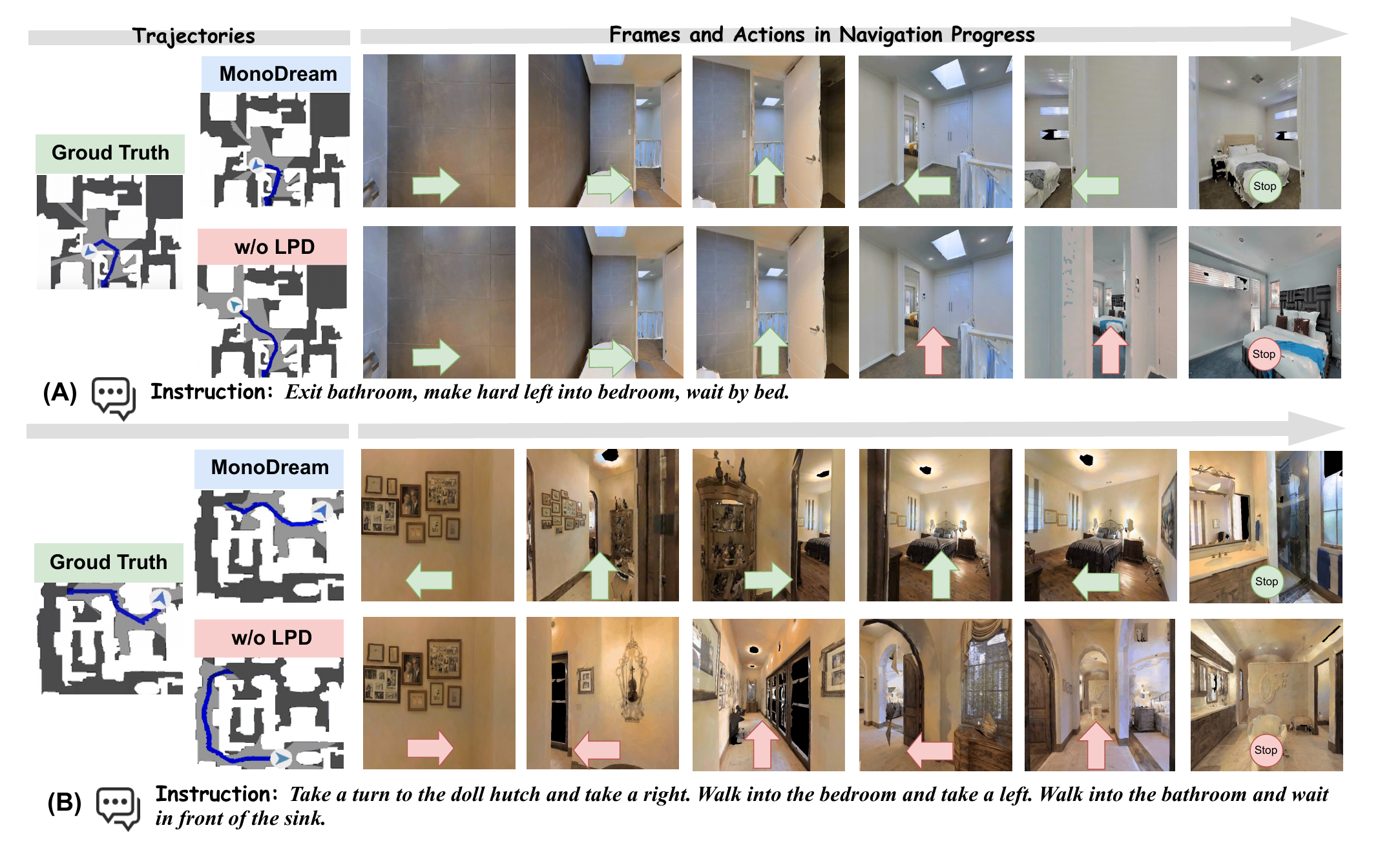}
    \caption{Quantitative results of MonoDream. We compare MonoDream with the ablated variant w/o LPD. Green arrows indicate correct actions, and red arrows indicate errors. (A) MonoDream correctly identifies the hard turning point at the fourth frame. In contrast, the w/o LPD baseline misreads the hallway layout, proceeds straight, and stops in the wrong room. (B) The w/o LPD model makes a critical mistake at the very first step, while MonoDream by leveraging internalized global features from LPD, correctly turns left even without explicit corner information in the initial monocular view. }
    \label{fig:qualitative}
\end{figure*}

\subsection{Latent Panoramic Dreaming}
\label{sec:lpd}

To supervise the learning of the proposed UNR $h_t$, we introduce a set of auxiliary tasks collectively referred to as Latent Panoramic Dreaming. These tasks guide the agent to enrich $h_t$ with the latent feature of the global scene, depth geometry, and future information, all without requiring explicit scene reconstruction.

First, we supervise the agent to align $h_t$ with the latent feature of the global and geometric context of the current scene from its limited forward-facing view. This encourages the learned UNR $h_t$ to capture a more complete spatial understanding beyond the monocular observation.

\begin{itemize}
    \item $\mathcal{H}^{\text{PI}}_t$: \text{Panoramic RGB at step } $t$.
    \item $\mathcal{H}^{\text{PD}}_t$: \text{Panoramic depth at step } $t$.
\end{itemize}

In addition, we guide the agent to anticipate how the scene evolves by predicting the latent features of the next-step panoramic RGB and depth, i.e. $\mathcal{H}^{\text{PI}}_{t+1}$ and $\mathcal{H}^{\text{PD}}_{t+1}$. This future-aware supervision encourages $h_t$ to incorporate short-term temporal dynamics and improves the agent's ability to plan ahead during navigation.

These latent supervision signals are obtained by encoding the corresponding panoramic RGB and depth using a vision encoder that shares weights with the visual encoder $\Phi_{\text{vis}}$, which is also used to encode the monocular input images.
This unified encoder design ensures that the supervision features and the learned UNR $h_t$ align in the same feature space and benefit from joint training.

The LPD training objective minimizes the mean squared error (MSE) between $h_t$ and the above latent targets. The feature loss function is:
\begin{equation}
L^{\text{Fea}}_{t}(\theta) = \sum_{m \in \mathcal{M}} \left\| h_t - \mathcal{H}_t^m \right\|^2 + \sum_{m \in \mathcal{M}} \left\| h_t - \mathcal{H}_{t+1}^m \right\|^2
\end{equation}
where $\mathcal{M} = \{\text{PI}, \text{PD}\}$.

Through this auxiliary supervision, LPD encourages the Unified Navigation Representation $h_t$ to internalize implicit global semantics, geometric layout, and short-term future cues. Our design improves the agent’s holistic understanding of the environment, leading to more informed and anticipatory navigation decisions.

Notably, the LPD module is used only during training. Panoramic RGB and depth signals are not required during inference, where the agent predicts actions based on monocular RGB images with the internalized global awareness.

\subsection{Multi-Task Co-Training}
\label{sec:cotrain}

In addition to the LPD task, which aligns the UNR $h_t$ with visual features, we further supervise $h_t$ using two language-based objectives: action prediction and instruction reasoning. These tasks align $h_t$ with the linguistic representation of navigation behaviors and goals, ensuring the learned UNR effectively bridges both visual and textual modalities.

At each time step $t$, the model predicts navigation actions in the natural language form. 
Specially, the model predicts a sequence of the next three actions $(a_t, a_{t+1}, a_{t+2})$ based on the embedding $h_t$, generated from the instruction $I$, current observation $o_t$, and navigation history $\mathcal{O}_t$ for the sample at step $t$. This setup encourages short-term action forecasting while retaining reactivity to new observations. The training objective for each timestamp $t$ is defined as:
\begin{equation}
L^{Act}_{t}(\theta) = - \sum_{k=0}^{K} \log \pi_{\theta}(a_{t+k}^* | h_t)
\end{equation}
where $a_{t+k}^*$ denotes the ground-truth action at step $t+k$.

To further enhance the agent’s understanding of instructions, we introduce instruction reasoning as another auxiliary task, which infers the underlying language instruction $\mathcal{I}$ based on the agent’s visual trajectory context, effectively promoting multimodal alignment from vision to language.

For each training trajectory, we uniformly sample $N$ visual observations to construct the input sequence $O_t$. These images are processed by the MonoDream backbone to produce the UNR $h_t$. A text decoder is then applied to generate the distribution over instruction tokens based on $h_T$, and the loss is computed accordingly:

\begin{equation}
    L^{Ins}_{\tau}(\theta) = - \log \pi_{\theta}(I_\tau | h_T)
\end{equation}

\begin{table*}[htbp]
  \centering
  \resizebox{1\textwidth}{!}{
  \begin{tabular}{l ccc c c c c r}
    \toprule
    \multirow{2}{*}{Method}  & \multicolumn{3}{c}{Observation} & \multicolumn{4}{c}{R2R Val-Unseen} & {Training} \\ 
    \cmidrule(lr){2-4} \cmidrule(lr){5-8} \cmidrule(lr){9-9}
     & {S.RGB} & {Depth} & {Pano.} & {NE $\downarrow$} & {OSR $\uparrow$} & {SR $\uparrow$} & {SPL $\uparrow$} & {External Data} \\
    \midrule
    BEVBert$^{\dagger}$\cite{an2022bevbert}        & & $\checkmark$ & $\checkmark$  & 4.57 & 67.0 & 59.0 & 50.0 &   - \\
    ETPNav$^{\dagger}$\cite{an2024etpnav}         & & $\checkmark$  & $\checkmark$  & 4.71 & 65.0 & 57.0 & 49.0 &   - \\
    ENP–ETPNav$^{\dagger}$\cite{liu2024vision}  & & $\checkmark$ & $\checkmark$  & 4.69 & {65}   & {58}   & {50}   &  - \\ 
    \midrule
    Seq2Seq$^\dagger$\cite{krantz2020beyond}      &  $\checkmark$ &$\checkmark$ & & 7.77 & 37.0 & 25.0 & 22.0 &  -  \\
    CMA$^\dagger$\cite{krantz2020beyond}      & $\checkmark$ &$\checkmark$ & & 7.37 & 40.0 & 32.0 & 30.0 &  -  \\
    LAW$^\dagger$\cite{ray2021language}    & $\checkmark$  &$\checkmark$ & & 6.83 & 44.0 & 35.0 & 31.0 &  -  \\
    CM2$^\dagger$\cite{georgakis2022cross}   & $\checkmark$ &$\checkmark$ & & 7.02 & 41.0 & 34.0 & 27.0 &  -  \\
    WS-MGMap$^\dagger$\cite{chen2022weakly}  & $\checkmark$ &$\checkmark$ & & 6.28 & 47.0 & 38.0 & 34.0 &   - \\
    sim2real$^\dagger$\cite{wang2024sim}   & $\checkmark$ & $\checkmark$ & & 5.95 & 55.8 & 44.9 & 30.4 &  -  \\ 
    NavMorph$^\dagger$\cite{yao2025navmorph} & $\checkmark$ & $\checkmark$ & & 5.75 & 56.9 & 47.9 & 33.2 &  -  \\ 
    NaVid-4D \cite{liunavid}     & $\checkmark$ & $\checkmark$ & & 5.99 & 55.7 & 43.8 & 37.1 & 1500K  \\
    \midrule
    NaVid \cite{zhang2024navid}     & $\checkmark$ & & & 5.47 & 49.1 & 37.4 & 35.9 &   \textbf{0K}  \\
    Uni-NaVid\cite{zhang2024uni}    & $\checkmark$ & & & 5.58 & 53.3 & 47.0 & 42.7 &  2300K  \\
    NaVILA\cite{cheng2024navila}    & $\checkmark$ & & & \textbf{5.22} & \underline{62.5} & 54.0 & \underline{49.0} & 2215K   \\
    Aux-Think \cite{wang2025think} & $\checkmark$ & & & 5.49 & \textbf{62.9} & \underline{55.7} & 48.7 & 1600K \\
    MonoDream (Ours) & $\checkmark$ & & & \underline{5.45} & 61.5 & \textbf{55.8} & \textbf{49.1} & \textbf{0K} \\
    \bottomrule
  \end{tabular}}
  \caption{Comparison of different methods on the R2R-CE Val-Unseen split. Observations used include single RGB camera (S.RGB), depth sensor (Depth) and panoramic view (Pano.). $\dagger$ indicates methods without using LLMs. External data refers to sources beyond the navigation simulator, such as real-world web data, general VQA datasets, and other similar resources.}
  \label{tab:r2r_val_unseen}
\end{table*}

During training, we co-train the action prediction and all the auxiliary tasks and switch between different tasks by changing the prompt (see our supplementary material for the prompts of different tasks). The final loss function is:
\begin{equation}
    L = \sum_{\tau \in D}( \sum_t^{T_\tau}(L^{Act}_{t}(\theta) + \lambda L^{Fea}_{t}(\theta)) + L^{Ins}_{\tau}(\theta))
\end{equation}
where $D$ is the set of training trajectories, $T_\tau$ is the step number of trajectory $\tau$, and $\lambda$ is the hyperparameter of weight.

\section{Implementation Details}

\subsubsection{Action Design} 
The action space of the agent is designed into four categories: move forward, turn left, turn right, and stop. The forward action includes step sizes of 25 cm, 50 cm, and 75 cm, while the turn actions are parameterized by rotation angles of 15$^\circ$, 30$^\circ$, and 45$^\circ$. This fine-grained design allows for more precise and flexible control, which is critical in complex environments.

\subsubsection{Training Datasets}

All the training data used in our work are collected from simulated environments, including the training splits of R2R-CE \cite{krantz2020beyond} and RxR-CE \cite{ku2020room}, as well as additional data collected using the DAgger \cite{ross2011reduction} strategy.
We first construct step-wise navigation data based on the action annotations provided in R2R-CE and RxR-CE, resulting in 320K and 600K samples respectively.
In addition, we construct auxiliary supervision data based on R2R-CE by applying the aforementioned image preprocessing and instruction reasoning tasks.
Moreover, following the DAgger strategy \cite{ross2011reduction}, we further collect 500K step-wise samples from non-oracle trajectories generated in the R2R-CE training environments.

For the training panoramic images in LPD tasks, we adapt the cubemap \cite{trindade2011improving} format and split into four canonical directions (left, front, right, and back). For the depth image, we apply log scaling to the raw values to compress large-scale variations, followed by a colormap-based RGB rendering \cite{opencv_library}.

\subsubsection{Model Training} We adopt NVILA-lite-2B \cite{liu2024nvila} as our base model, which includes a SigLIP vision encoder \cite{tschannen2025siglip}, a projection module, and a Qwen2-based language model \cite{bai2025qwen2}. Starting from NVILA-lite-2B's pre-trained weights, we perform supervised fine-tuning for our tasks. All components of the model are trainable during fine-tuning.
Training is conducted on 8 NVIDIA H20 GPUs for 5 epochs with the learning rate of 1e-5 , the warm-up as 0.03, and the batch size of 80.
During training and inference, we set the number of future actions to predict ($K$) to 3, and the number of historical frames to sample ($N$) to 8.

\section{Experimental Results}
\label{sec::experimental}

\subsection{Experiment Setup}
\label{sec:exp_set}

\subsubsection{Simulated environments} 
We evaluate our method on the VLN-CE benchmarks R2R-CE \cite{krantz2020beyond} and RxR-CE \cite{ku2020room} following the standard VLN-CE settings. 
All the methods are evaluated on the R2R val-unseen split and RxR val-unseen split. The quantitative results are in Figure \ref{fig:qualitative}.

\subsubsection{Metrics} We follow the standard VLN evaluation protocol \cite{krantz2020beyond, ku2020room} to evaluate navigation performance for all methods, including success rate (SR), oracle success rate (OSR), success weighted by path length (SPL), and the navigation error from the goal (NE). Among them, SR and SPL are widely regarded as the primary metrics, reflecting the task completion and path efficiency respectively.

\subsection{Comparison on VLN-CE Benchmarks}

We evaluate our method on the VLN-CE benchmarks inluding R2R-CE and RxR-CE, which provide continuous environments for navigational actions in reconstructed photorealistic indoor scenes.
We first focus on the val-unseen split in R2R-CE dataset in Table \ref{tab:r2r_val_unseen}.
To ensure a fair comparison, we group methods based on their sensor settings and mark those that do not rely on large language models ($\dagger$).

As in Table \ref{tab:r2r_val_unseen}, our proposed MonoDream achieves strong performance without relying on any external data beyond the simulation datasets.
This demonstrates the data efficiency of our method, which learns effective navigation behaviors. We attribute this success primarily to the incorporation of LPD supervision. By jointly training on panoramic images, panoramic depth maps, and future panoramic views, the model learns to build a more comprehensive understanding of the scene structure and spatial layout. This enhances the model's generalization ability, even when trained solely on limited simulator-generated data. The synergy between action prediction and LPD further amplifies the effectiveness of each training sample, enabling the model to extract richer visual cues and improve data efficiency.

To further evaluate MonoDream's robustness and long-term ability, we conduct experiments on the RxR-CE Val-Unseen split \cite{ku2020room}, as in Table~\ref{tab:app_rxr_val_unseen}. Compared to R2R-CE, RxR-CE presents significantly more complex challenges due to its longer trajectories and more natural, diverse language instructions. Despite these challenges and the monocular setting, MonoDream achieves state-of-the-art results on the primary metrics (SR and SPL), outperforming strong baselines like Uni-NaVid \cite{zhang2024uni} and NaVILA \cite{cheng2024navila} while using substantially less training data. These findings demonstrate the advantage of leveraging  visual imagination as auxiliary supervision.

\subsubsection{Cross-dataset Evaluation}

We assess the generalization capability of MonoDream on the RxR-CE Val-Unseen split (Table~\ref{tab:rxr_results}). Notably, all the methods are trained without any RxR-CE training data and our method achieves the state-of-the-art performance.

The effectiveness of our model in a cross-dataset setting highlights three key insights. First, the use of LPD strengthens the backbone’s ability to form a richer and more generalizable representation of the environment. Second, our LPD tasks exhibit strong generalization capabilities, with their benefits extending beyond the distribution of the training set. Third, for long-horizon navigation tasks in RxR-CE, having a global understanding or imagination of the environment significantly enhances the agent's navigation capabilities.

\begin{table}[htbp]
  \centering
  \scalebox{0.88}{
  \begin{tabular}{l c c c c}
    \toprule
    \multirow{2}{*}{Method} & \multicolumn{4}{c}{RxR Val-Unseen} \\ 
     \cmidrule(lr){2-5} 
     & {NE $\downarrow$} & {OSR $\uparrow$} & {SR $\uparrow$} & {SPL $\uparrow$}  \\
    \midrule
    CMA$^\dagger$ \cite{hong2022bridging}           & 8.76 & - & 26.5 & 22.1 \\
    VLNBERT$^\dagger$ \cite{hong2022bridging}           & 8.98 & - & 27.0 & 22.6 \\
    Seq2Seq$^\dagger$\cite{krantz2020beyond}   & 11.8& - & 13.9 & 11.9  \\
    sim2real$^\dagger$\cite{wang2024sim}   & 8.79 & 36.7 & 25.5 & 18.1   \\
    Uni-NaVid\cite{zhang2024uni}   & \textbf{6.24} & \underline{55.5} & 48.7 & \underline{40.9}  \\
    NaVILA\cite{cheng2024navila}   & {6.77} & - & \underline{49.3} & \textbf{44.0}   \\
    MonoDream (Ours)   & \underline{6.38} & \textbf{55.8}  & \textbf{49.4} & \underline{40.9}   \\
    \bottomrule
  \end{tabular}}
  \caption{Comparison of different methods on the RxR Val-Unseen split. $\dagger$ indicates methods without using LLMs.}
  \label{tab:app_rxr_val_unseen}
\end{table}

\begin{table}[htbp]
\centering
\scalebox{0.9}{
\begin{tabular}{@{}l c c c c@{}} 
\toprule
\multirow{2}{*}{Method} & \multicolumn{4}{c}{RxR Val-Unseen} \\
 \cmidrule(lr){2-5}
& {NE $\downarrow$} & {OSR $\uparrow$} & {SR $\uparrow$} & {SPL $\uparrow$} \\
\midrule
Seq2Seq\cite{krantz2020beyond}  & 11.8            & 5.02            & 3.51            & 3.43            \\
CMA\cite{krantz2020beyond}      & 11.7            & 10.7            & 4.41            & 2.47            \\
LAW\cite{ray2021language} & 10.87           & 21.0            & 8.0             & 8.0             \\
CM2\cite{georgakis2022cross}   & 8.98            & 25.3            & 14.4            & 9.2             \\
WS-MGMap\cite{chen2022weakly} & 9.83            & 29.8            & 15.0            & 12.1            \\
A$^2$NAV\cite{chen20232}     & {-}             & {-}             & 16.8            & 6.3             \\ 
NaVid\cite{zhang2024navid}   & \textbf{8.41}   & \underline{34.5}  & \underline{23.8}  & \underline{21.2}  \\
MonoDream (Ours)  & \underline{8.57}  &  \textbf{35.9}  & \textbf{25.1}   &  \textbf{21.6} \\
\bottomrule
\end{tabular}}
\caption{Cross-dataset performance on the RxR-CE Val-Unseen split. All results are obtained without training on the RxR-CE training set. }
\label{tab:rxr_results}
\end{table}

\subsection{Ablation Study}

In the ablation study, we separately evaluate the effectiveness of the two auxiliary tasks, LPD and Instruction Reasoning, as well as the individual contributions of the four sub-tasks in LPD. To ensure efficiency, all ablation experiments are conducted using only the R2R-CE training split and test on R2R-CE val-unseen split only.

\subsubsection{Impact of Auxiliary Tasks}

We conduct ablation experiments to evaluate the contribution of the two auxiliary tasks: Instruction Reasoning and LPD. As shown in Table~\ref{tab:ablation1}, removing both auxiliary tasks leads to the poorest performance. Introducing IR alone yields moderate gains across all metrics, demonstrating that aligning the UNR $h_t$ with language-based supervision (i.e., instruction prediction) enhances the model’s understanding of high-level navigation goals.

The most substantial improvement comes from LPD. Introducing LPD achieves the best performance across all metrics. These results confirm the complementary benefits of language- and vision-oriented supervision: IR improves semantic goal alignment, while LPD enriches UNR with global, geometric, and predictive spatial cues that are otherwise missing in monocular observations.

\begin{table}[htbp]
\centering
\scalebox{1}{
\begin{tabular}{cc|cccc}
\toprule
\multicolumn{2}{c|}{Auxiliary Tasks} & \multicolumn{4}{c}{Metrics} \\
 IR & LPD & NE$\downarrow$    & OSR$\uparrow$    & SR$\uparrow$     & SPL$\uparrow$ \\ \midrule
    &      & 7.78 & 43.7 & 35.1 & 30.2 \\
\checkmark & & \underline{7.67} & \underline{45.8} & \underline{37.7} & \underline{32.1} \\
 \checkmark &\checkmark &  \textbf{6.19}  &  \textbf{51.1} &  \textbf{46.1}   &  \textbf{39.9} \\
\bottomrule
\end{tabular}}
\caption{Ablation study on auxiliary tasks. IR: Instruction Reasoning; LPD: Latent Panoramic Dreaming}
\label{tab:ablation1}
\end{table}

\subsubsection{Impact of Auxiliary Panoramic Dreaming Tasks}

Table~\ref{tab:ablation} presents the ablation study of the four auxiliary tasks in LPD, including the latent feature of panoramic RGB image (PI), panoramic depth (PD),  future panoramic RGB image (FPI), and future panoramic depth (FPD). We add each component on top of the baseline and evaluate their individual and cumulative contributions to navigation performance.

The results show that each LPD task brings consistent improvements, confirming their effectiveness as auxiliary supervision signals. In particular, latent panoramic and depth provide the most significant gains. We attribute this to their role in enhancing the agent's spatial and structural understanding of the environment, while implicit depth introduces geometric awareness that complements RGB observations.

\begin{table}[htbp]
\centering
\scalebox{0.9}{
\begin{tabular}{cccc|cccc}
\toprule
\multicolumn{4}{c|}{LPD Tasks} & \multicolumn{4}{c}{Metrics} \\
PI & PD & FPI & FPD & NE$\downarrow$    & OSR$\uparrow$    & SR$\uparrow$     & SPL$\uparrow$ \\ \midrule
  &   &    &     & 7.67             & {45.8} & 37.7             & 32.1 \\
\checkmark & & & & 7.22             & 44.2             & 39.6             & 35.3 \\
 &\checkmark & & & \underline{6.71}    & \underline{47.4}    & \underline{42.2}    & \underline{37.7} \\
 & &\checkmark & & 7.03             & 44.5             & 39.1             & 34.2 \\
 & & &\checkmark & 6.80 & 45.3             & 39.8 & 35.4 \\
 \checkmark & \checkmark & \checkmark & \checkmark & \textbf{6.19}  &  \textbf{51.1} &  \textbf{46.1}   &  \textbf{39.9} \\
\bottomrule
\end{tabular}}
\caption{Ablation study on four LPD tasks. PI: Panoramic RGB image; PD: Panoramic depth; FPI: Future panoramic RGB image; FPD: Future panoramic depth.}
\label{tab:ablation}
\end{table}

\subsubsection{Impact of Prediction Steps in Latent Panoramic Dreaming}

We further investigate the effect of predicting the future panoramic feature of different future steps in LPD, and we take FPI as a representative example. As shown in Table~\ref{tab:ablation_fpi}, predicting a single future step delivers the best performance. Extending the prediction horizon to 2 or 3 future steps leads to performance degradation, due to compounded uncertainty in long-range forecasting under partial observations.

\begin{table}[htbp]
\centering
\scalebox{0.95}{
\begin{tabular}{c|cccc}
\toprule
\#step & NE$\downarrow$ & OSR$\uparrow$ & SR$\uparrow$ & SPL$\uparrow$ \\
\midrule
1  & \textbf{7.03}   & \textbf{44.5}  & \textbf{39.1}  & \textbf{34.2} \\
2  & \underline{7.89} & \underline{43.9} & \underline{34.3} & \underline{29.1} \\
3  & 8.77 & 40.0 & 29.2 & 26.2 \\
\bottomrule
\end{tabular}}
\caption{Ablation on the prediction steps in LPD.}
\label{tab:ablation_fpi}
\end{table}

\subsection{Model Efficiency}

In addition to navigation performance, we compare the efficiency of our method MonoDream with recent state-of-the-art VLN methods. Specifically, we report the size of model parameters and the average inference time per step on a single NVIDIA 4090 GPU with the same local hardware.

As in Table~\ref{tab:efficiency}, our method achieves smaller model size and fastest inference speed among compared approaches. Despite introducing auxiliary latent imagination tasks during training, MonoDream is inference-efficient as auxiliary modules are disabled at test time. This makes MonoDream a practical solution for real-time embodied navigation applications.

\begin{table}[h]
\centering
\scalebox{0.95}{
\begin{tabular}{lcc}
\toprule
Method & Params. &  Time / Step \\
\midrule
NaVILA \cite{cheng2024navila}           & 8B          & 1.2s              \\
Aux-Think \cite{wang2025think} & 8B & 1.2s\\
MonoDream (Ours) & \textbf{2B} & \textbf{0.8s} \\
\bottomrule
\end{tabular}}
\caption{Comparison of model efficiency.}
\label{tab:efficiency}
\end{table}

\section{Conclusion}

In this work, we present MonoDream, a unified framework for vision-language navigation (VLN) that leverages a vision-language model as the backbone to integrate action prediction and latent imagination. 
We propose the Unified Navigation Representation to align all navigation-relevant information and use LPD task to co-train the UNR.
MonoDream achieves SOTA performance on the monocular VLN-CE benchmark by implicitly modeling panoramic understanding, depth perception, and future view prediction by latent supervision. 

Our model training is conducted using only simulator-based data from the R2R-CE and RxR-CE datasets, without relying on external data. Our experiments demonstrate that MonoDream can achieve strong performance using only monocular inputs, narrowing the gap with methods that explicitly consume panoramic or depth data. 
This work highlights the potential of implicit multimodal learning as a scalable and efficient solution for embodied navigation agents using only monocular inputs.

\textbf{Limitations.} 
MonoDream imagines the current scene and immediate future from past monocular observations, without explicitly reconstructing panoramic history. 
Incorporating richer temporal modeling, such as  memory-based reasoning, may further improve planning and robustness.


\bibliography{aaai2026}

\newpage

\section{Appendices}

\subsection{Prompts for Different Tasks}

We use the following prompt to drive the model to predict navigation actions:
\begin{tcolorbox}[
  colback=gray!15,    
  colframe=black!75,  
  boxrule=0.5pt,      
  arc=1mm,            
]
Imagine you are a robot programmed for navigation tasks. You have been given a video of historical observations: <image>,...,<image> and and current observation: <image>. Your assigned task is: [Instruction]. Analyze this series of images to decide your next move, which could involve turning left or right by a specific degree, moving forward a certain distance, or stop if the task is completed.
\end{tcolorbox}

Among them, [Instruction] is the language instruction given for the current task. 
For the instruction reasoning, the prompt is set to be:
\begin{tcolorbox}[
  colback=gray!15,    
  colframe=black!75,  
  boxrule=0.5pt,      
  arc=1mm,            
]
Imagine you are a robot designed for navigation. You are provided with captured image sequences: <image>,...,<image>. Based on this image sequence, please describe the navigation trajectory of the robot.
\end{tcolorbox}

For the LPD tasks, we design four variants: predicting the latent features of the current panoramic RGB image, the current panoramic depth map, the future panoramic RGB image, and the future panoramic depth map. The prompts for these tasks are respectively set as:
\begin{tcolorbox}[
  colback=gray!15,    
  colframe=black!75,  
  boxrule=0.5pt,      
  arc=1mm,            
]
Imagine you are a robot programmed for navigation tasks. You have been given a video of historical observations: <image>,...,<image> and and current observation: <image>.  Analyze the series of images to predict the panoramic image of current observation.
\end{tcolorbox}

\begin{tcolorbox}[
  colback=gray!15,    
  colframe=black!75,  
  boxrule=0.5pt,      
  arc=1mm,            
]
Imagine you are a robot programmed for navigation tasks. You have been given a video of historical observations: <image>,...,<image> and and current observation: <image>.  Analyze the series of images to predict the pamoramic depth of current observation.
\end{tcolorbox}

\begin{tcolorbox}[
  colback=gray!15,    
  colframe=black!75,  
  boxrule=0.5pt,      
  arc=1mm,            
]
Imagine you are a robot programmed for navigation tasks. You have been given a video of historical observations: <image>,...,<image> and and current observation: <image>. Your assigned task is: [Instruction]. Analyze the series of images to predict the panoramic image of current observation.
\end{tcolorbox}

\begin{tcolorbox}[
  colback=gray!15,    
  colframe=black!75,  
  boxrule=0.5pt,      
  arc=1mm,            
]
Imagine you are a robot programmed for navigation tasks. You have been given a video of historical observations: <image>,...,<image> and and current observation: <image>. Your assigned task is: [Instruction]. Analyze the series of images to predict the panoramic depth of current observation.
\end{tcolorbox}

\subsection{Supervision Image Data Preprocessing for LPD}

For panoramic RGB images, we adapt the cubemap \cite{trindade2011improving} format and discretize the 360-degree field of view into four canonical directions (left, front, right, and back) each represented by a 90-degree field-of-view sub-image. These four perspective views are as the panoramic observation and serve as input to the vision encoder. This decomposition improves alignment with the inductive biases of standard vision models, which are typically pretrained on perspective images rather than 360-degree panoramas.

For depth maps, we first apply logarithmic scaling to the raw depth values to compress large-scale variations, followed by a colormap-based RGB rendering \cite{opencv_library}. The resulting pseudo-RGB depth image is then compatible with standard pretrained encoders. This log-depth encoding enhances the visual semantics and makes the depth modality more accessible to the model without requiring specialized depth pretraining.

These preprocessing steps are motivated by the fact that most existing vision encoders, such as SigLIP \cite{tschannen2025siglip}, are pretrained on natural perspective RGB images. By transforming panoramic and depth modalities into visually aligned formats, we maximize compatibility with the pretrained knowledge of the vision backbone. We verify the effectiveness of our design in the ablation experiments.

\subsection{Ablation Study on Different Image Pre-processing Methods for LPD}
\label{sec:ab_data}

Most existing vision-language models are pretrained on general RGB images and thus lack exposure to raw panoramic or depth representations. Moreover, since large language models (LLMs) are trained to predict text embeddings rather than image embeddings, a distribution gap exists between modalities. Effectively leveraging visual pretraining becomes essential for narrowing this gap.
To investigate the image format with minimal cost, we conduct ablation studies using only the training split of R2R-CE with only one epoch training.

For panoramic images, we compare two representations: a single equirectangular image with a 360-degree field of view, and a cubemap composed of four 90-degree field-of-view images (left, front, right, back). Our results in Table \ref{tab:ablation_pano} show that the cubemap format performs better, likely because its visual structure better aligns with the distribution of pretrained vision encoders.

\begin{table}[htbp]
\caption{Ablation study on the panoramic image formats.}
\centering
\begin{tabular}{c|cccc}
\toprule
Pano. Format    & NE$\downarrow$ & OSR$\uparrow$ & SR$\uparrow$ & SPL$\uparrow$  \\ \midrule
Equirectangular & 7.53 & \textbf{46.6} & 38.5 & 33.2 \\
Cubemap         & \textbf{7.22} & 44.2 & \textbf{39.6} & \textbf{35.3} \\ \bottomrule
\end{tabular}
\label{tab:ablation_pano}
\end{table}

For depth images, we convert depth values into RGB representations. We compare three preprocessing strategies: inverse depth, log-transformed depth, and linear normalization of depth values. Among them, log-transformed depth achieves the best performance, possibly because human depth perception is inherently nonlinear. The logarithmic  distribution helps the model focus more on semantically meaningful depth ranges. The results are shown in Table \ref{tab:ablation_depth}.

\begin{table}[htbp]
\caption{Ablation study on the panoramic depth formats and normalization methods.}
\centering
\scalebox{0.85}{
\begin{tabular}{cc|cccc}
\toprule
Format  & Norm.  & NE$\downarrow$ & OSR$\uparrow$ & SR$\uparrow$ & SPL$\uparrow$  \\ \midrule
Cubemap & Linear      & 6.92 & 46.3 & 42.2 & 37.1 \\
Equirectangular & Logarithmic & 7.07 & 45.1 & 39.9 & 35.1 \\
Cubemap & Logarithmic & \textbf{6.71} & \textbf{47.4} & \textbf{42.2} & \textbf{37.7} \\ \bottomrule
\end{tabular}}
\label{tab:ablation_depth}
\end{table}

\end{document}